# Mobile Robot Path Planning in Dynamic Environments: A Survey

Kuanqi Cai [1,2], Chaoqun Wang [2], Jiyu Cheng[2], Shuang Song[1],
Clarence W. de Silva[1,3], *Fellow, IEEE*, and Max Q.-H. Meng[1, 2], *Fellow, IEEE*

(1. School of Mechanical Engineering and Automation, Harbin Institute of Technology (Shenzhen), Shenzhen 518055;
2. Department of Electronic Engineering, The Chinese University of Hong Kong, Hong Kong SAR 999077;
3. The University of British Columbia, Vancouver, BC, Canada;

**Abstract:** There are many challenges for robot navigation in densely populated dynamic environments. This paper presents a survey of the path planning methods for robot navigation in dense environments. Particularly, the path planning in the navigation framework of mobile robots is composed of global path planning and local path planning, with regard to the planning scope and the executability. Within this framework, the recent progress of the path planning methods is presented in the paper, while examining their strengths and weaknesses. Notably, the recently developed Velocity Obstacle method and its variants that serve as the local planner are analyzed comprehensively. Moreover, as a model-free method that is widely used in current robot applications, the reinforcement learning-based path planning algorithms are detailed in this paper.

## 1 Introduction

Autonomous robots, such as industrial robots in factories and service robots in public areas, have attracted much attention and increasingly developed in the past few decades. For example, the Multi-Modal Mall Entertainment Robot (MuMMER) developed in [1] can provide the required service to the public in an open-air plaza environment. Atlas robot [2] developed by Boston Dynamics can perform special tasks in both indoor and outdoor environments. More recently, Tanaka [3] developed a human-like robot called Pepper, which can serve as a family companion to assist elder people in domestic environments. A trend exists that these robots are increasingly used for or co-existing with humans in more and more complex environments, such as shopping malls, city streets, or train stations. In these environments, it is a prerequisite that the robot plans a safe and collision-free path for providing services to humans in an effective manner.

In recent years, many review articles about robot path planning have been published. For instance, Kruse et al. [4] surveyed the topic of socially-aware trajectory planning. This article focused more on robot behavior during the navigation, in which factors such as the human comfort and the sociality are investigated in path planning. Chik et al. [5] divided path planners for robot navigation into a global planner and a local planner. Their survey outlines several types of framework for robot navigation but lacks adequate description of the local path planner. Douthwaite et al. [6] presented a comparative study of Velocity Obstacle (VO) methods for multiple agents. They proposed as well several evaluation metrics to cope with the uncertainty caused by low- resolution sensors in the robot. Mohanan et al. [7] reviewed the research on robot motion planning in complicated environments. They classified motion planning methods and carried out a comparative study of their performance. Zafar et al. [8] divided the motion planning methods into typical methods and heuristic methods. The comparative analysis of the two categories showed that the heuristic approach exhibited better performance in path planning. More recently, Cheng et al. [9] divided the current methods into reactive-based methods, predictive-based methods, model-based methods, and learning-based methods. These surveys are outlined in Table 1, chronologically, and the main features of these methods are highlighted.

Although many available surveys introduce path planning in robot navigation in different aspects, there is not a comprehensive study that systematically introduces the hierarchical path planning framework involving the global path planner and local path planner. Moreover, the VO method, as a local path planner for obstacle avoidance, is not well investigated. To address these weaknesses, the present paper introduces in detail different methods that can be served as the global and the local path planner. Furthermore, the discussion is extended from the model-based methods into the learning-based framework. The Reinforcement Learning (RL) method, which is able to solve the path planning problem comprehensively, is investigated in the present paper. These methods are outlined and their recent progress in solving the path planning problem is presented.

Table 1 Summary of The Literature on Robot Obstacle Avoidance

| Author | Year | Main content |
|---|---|---|
| **Thibault Kruse** [4] | 2013 | Factors that robots need to concern in the navigation |
| **S. F. Chik** [5] | 2016 | Global planner, local planner, four types of navigation framework |
| **Douthwaite J.A** [6] | 2018 | Comparative study of velocity obstacle methods |
| **M.G.Mohanan** [7] | 2018 | Robot motion planning |
| **MN Zafar** [8] | 2018 | Classical approach, Heuristic approach |
| **Jiyu Cheng** [9] | 2018 | Reactive-based methods, Predictive-based methods, Model-based method, and Learning-based methods |

The rest of the paper is organized as follows. Section 2 introduces the classical framework of navigation. The state-of-the-art of classical global planners and traditional methods for local path planning that focus on VO are introduced. Reinforcement learning methods for path planning are introduced in Section 3. Conclusions of the paper are given in Section 4.

## 2 Hierarchical Path Planning

The path planning module plays a key role in guiding the robot to move safely in dynamic environments. The objective of path planning is to guide the robot to move from the initial point to the target point, subject to the vehicle motion constraints. Generally, path planning can be separated into two stages [10]: global planner and local planner, based on the environmental information that the robot can acquire during the navigation process.

The navigation framework of a robot consists of a global path planner and a local path planner, as shown in Fig. 1. Formally, the global path planner accounts for planning a collision-free path from the starting point to the goal point. Only a static global map is involved in this process, thus the generated global path is static, without considering the dynamic objects. The local path planner proceeds to piecewise optimization of the global path, which takes into consideration the information of dynamic objects and robot motion constraints. Adopting this hierarchical path planning framework results in many benefits in real applications. The recent developments of global path planning algorithms and local path planning algorithms are reviewed in the following sections.

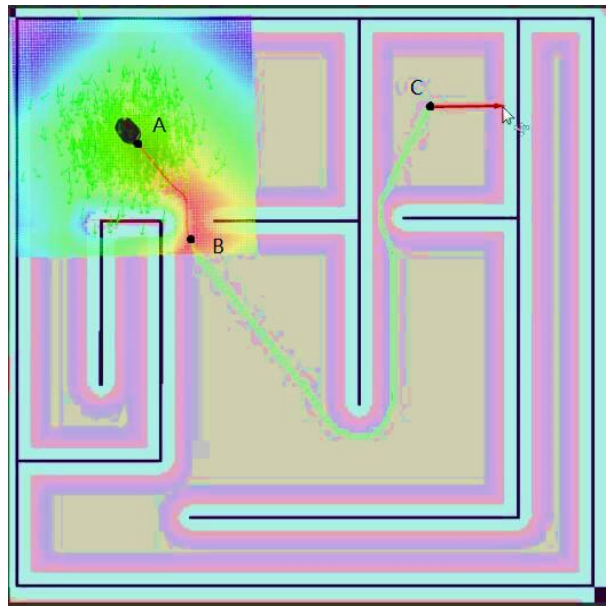

Fig. 1 The paths planned by the global planner and local planner. Start point is marked as point A. The target of robot is point C. The green path is produced by the global planner while the red track is produced by the local planner.

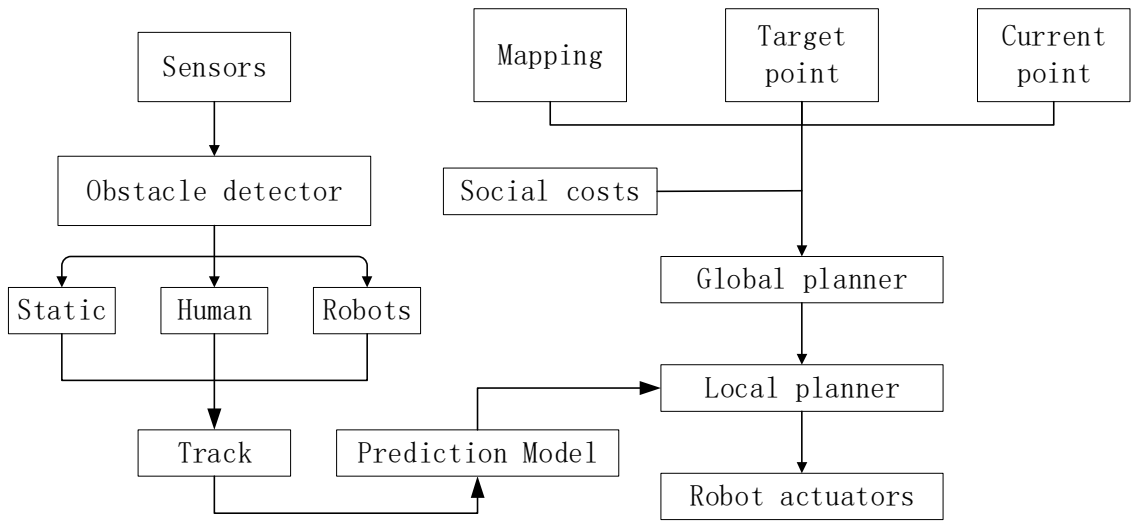

Fig. 2 Navigation framework having a global path planner and a local path planner.

## 2.1 Global path planner

The global path planning method generates a path for the robot according to the global map and the target point, It is necessary to plan the path that the robot can follow by using the map information obtained in a dynamic environment, to cope with any obstacles that may appear at any time.

A large body of research is available on global path planning. The developed algorithms can be divided into three categories: graph search-based algorithm [11], random sampling algorithm [12], and intelligent bionic algorithm [13], as shown in Fig. 3. Classical graph-search-based algorithms for graph search mainly include the Dijkstra algorithm [14], A* algorithm [15], DFS algorithm [16], and BFS algorithm [17]. Dijkstra algorithm and A* algorithm are well investigated over the past few decades and have demonstrated their capability by widely being implemented with the Robot Operating System (ROS) [18] for real-world robot applications. With the heuristic searching strategy, these methods are effective in relatively simple 2D environments. However, these methods suffer from a heavy computational burden when implemented in large-scale or high-dimensional environments.

Generally, as shown in Fig. 3, random sampling algorithms include Batch Informed Trees (BIT) [19], Regionally Accelerated Batch Informed Trees (RABIT) [20], Rapidly-exploring Random Tree (RRT) [21], and Risk-based Dual-Tree Rapidly exploring Random Tree (Risk-DTRRT) [22], etc. Compared with the graph search-based algorithms, these algorithms are more efficient and widely used in dynamic or high-dimensional environments.

Another important branch of the global path planning methods is the intelligent bionic-based method, which is a type of intelligent algorithm that simulates the evolutionary behaviors of insects. It generally includes Genetic Algorithm (GA) [23], Ant Colony Algorithm (ACO) [24], Artificial Bee Colony Algorithm (ABC) [25], and Particle Swarm Optimization Algorithm (PSO) [26]. To further improve the calculation efficiency and avoid local optima problems, a lot of advanced algorithms are proposed. Wang et al. [27] proposed the Optimization of the Genetic Algorithm-Particle Swarm Optimization Algorithm (OGA-PSO) to solve the shortest collision-free path planning problem of a welding robot. Liu et al. [28] combined the artificial potential field and geometric local optimization method with ACO to search for the globally optimal path. Mac et al. [29] put forward a constrained multi-objective particle swarm optimization algorithm with an acceleration methodology to generate an optimal global trajectory.

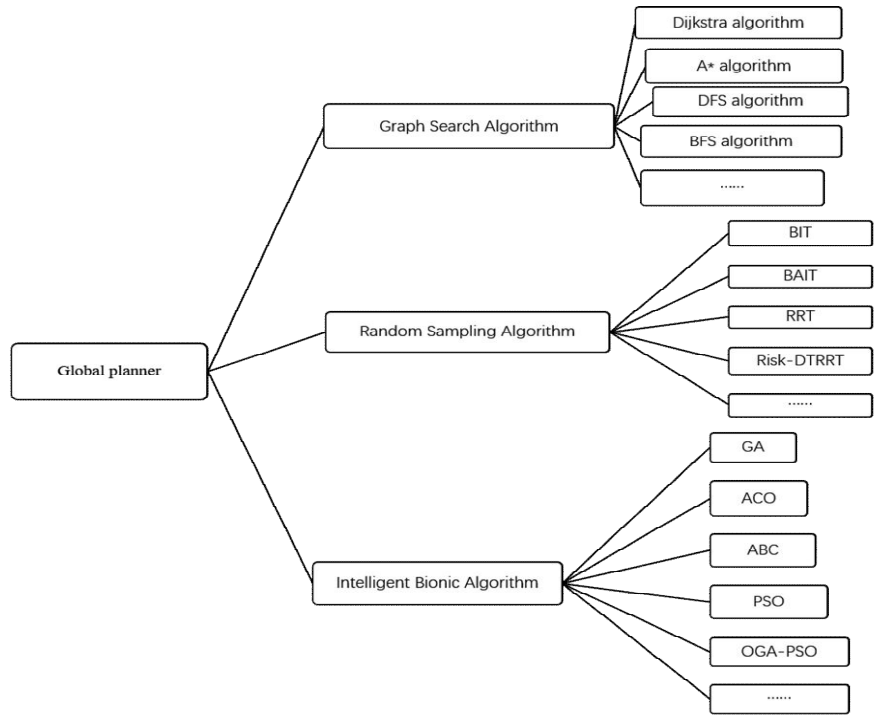

Fig. 3 Diagrammatic representation of different classical methods of global planner.

## 2.2 Local path planner

The local path planner focuses on generating a local path by using the available information on the surrounding environment of the robot so that the robot can avoid local obstacles in an effective manner. Local path planners are widely used because the information captured by the sensor system changes in real-time in dynamic environments. Compared with the global path planning method, the local path planning method is more efficient and practical, serving as the bridge between the global path and control. However, one notable disadvantage is that the local planner may be trapped in a local minimum.

Usually, humans are regarded as obstacles in robot navigation. Nishitani et al. [30] developed X-Y-T space motion planning method to avoid humans. It considers the human orientation and human personal domain. However, the calculation efficiency heavily depends on the grid size of the navigation map. Kollmitz et al. [31] proposed a layered social cost map for navigation in complex environments. Moreover, as an extension of A* algorithm [32], the timed A* method is able to predict human trajectories by using a social cost function.

There are many classical algorithms for obtaining an optimal path in the local planner and avoiding the local minima problem, such as Artificial Potential Field method [33], Fuzzy Logic Algorithm [34], Simulated Annealing Algorithm [35], Particle Algorithm [36], and hybrid method combined with the genetic algorithm [37]. However, these methods do not consider the relative motion between the agent and the dynamic objects, and even worse, sometimes it is hard to explicitly acquire the velocity profiles of the dynamic objects.

Recently, a local planner that does not rely on explicit velocity profile information has been developed. In this system, the navigation of each agent in the environment is independent, and an agent does not need to communicate with other agents [38]. Particularly, P. Fiorini [39] put forward the theory of VO by defining a velocity constraint described as a geometric region, into which the speed of the agent falls, which will cause a collision among agents at the next step. The method is effective in obstacle avoidance while concerning the agent velocity. However, with VO, the oscillation will occur when two agents are on the collision course with each other. These oscillations arise from the fact that both robots choose a larger offset of current velocity in the beginning of obstacle avoidance. To reduce the offset of the current velocity and improve the performance, Van den Berg et al. [38] came up with the reciprocal velocity obstacle (RVO) method. They regarded the new velocity profile of the robot as the average value of its current velocity and the velocity that lies out of other agents' VO. The RVO is suggested to be a useful way to plan a smooth and safe path without oscillations, in multiple-agent navigation. However, it still has a disadvantage that multiple robots may not be able to arrive at a consensus on which side to go through, which causes a problem called "reciprocal dance". To solve this problem, Snape [40] extended RVO to Hybrid Reciprocal Velocity Obstacle (HRVO). This approach has been applied for multi-robot navigation by considering the kinematics and sensor uncertainty of the robot.

Yet if there are numbers of dynamic objects in the application scenario, the robot velocity will approach to the initial point in the velocity space [41]. As a result, the robot may be trapped in one area. This problem can be eliminated by the truncation approach [42], with which robots will not collide at a defined timestep after truncation. An example of VO is shown in Fig. 4, where the gray areas mark the velocity profile with which may cause collision among agents. More details can be found in [43]. The new velocity of agents must be chosen outside these gray regions. To this end, multiple methods in different aspects are proposed. Three commonly used methods that have been proposed in recent years are outlined now. The first method is Optimal Reciprocal Collision Avoidance (ORCA) introduced by Berg et al. [44]. With this method, half-planes of collision-free velocities can be calculated and assigned for each agent. The optimal velocity region can then be defined by solving a linear program problem. Agents select a velocity profile that is closest to the optimal velocity and move with it next. The second method that estimates collision-free velocity is the ClearPath raised in [45]. ClearPath is a robust method and is better than prior VO-based methods for collision avoidance. There are two ways for ClearPath to calculate the collision-free velocity. One is to choose velocity at the intersection of two boundary lines of arbitrary VOs. Another method chooses the velocity determined by the projection of the preferred velocity profile onto the nearest VO [44]. The third method is the Collision Avoidance with Localization Uncertainty (CALU) method introduced in [46], which combines Optimal Reciprocal Collision Avoidance (ORCA), and Non-holonomic robots Optimal Reciprocal Collision Avoidance (NH-ORCA) [43] to alleviate the need for prior knowledge on the environment.

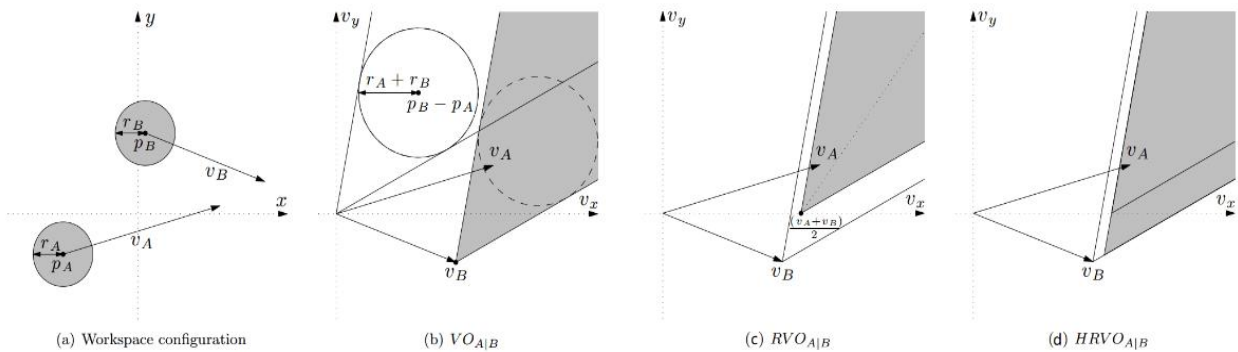

Fig. 4 A Sample figure about VOs. Figure courtesy [42].

VOs can effectively cope with the problems caused by inaccurate localization and communication in complex environments. Still, it needs adequate messages about the shape, velocity, and position of the agent. However, it is useless under two conditions: a) The robot chassis cannot be treated as a disc; b) The pose belief distribution of AMCL drifts in one direction. Claes [41] introduced Collision Avoidance under Bounded Localization Uncertainty (COCALU) to solve this problem. It changes the form of the particle cloud rather than the circumscribed circle. The evolution of these VOs is depicted in Fig. 5. Table 2 summarizes the available local path planners.

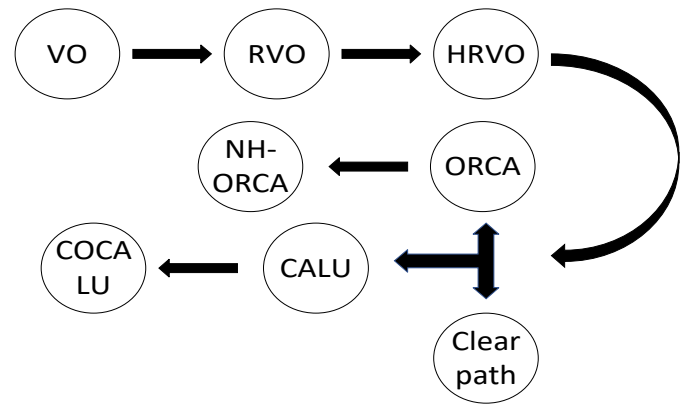

Fig. 5 Relationship of classical algorithms based on velocity obstacle.

Table 2 Classic algorithms of the local planner

| Local planners | Advantages | Disadvantages |
|---|---|---|
| **Artificial Potential Field method [33]** | The scheme has high efficiency and it can solve the local minimum problem in the traditional algorithm | There is a trapping area and the robot will oscillate when it passes through the narrow passage. |
| **Fuzzy Logic Algorithm [34]** | It reduces the dependence on environmental information and has the advantages of good robustness and effectiveness | Fuzzy rules are often predetermined by people's experience, so they are unable to learn and have poor flexibility |
| **Simulated Annealing Algorithm [35]** | Simple description, flexible use, high efficiency, less initial conditions | Slow convergence and high randomness |
| **VOs [39]** | It considers the obstacle velocity | The complex relationships between societies are not considered |
| **X-Y-T Space [30]** | It considers the human's directional area and personal space | Efficiency depends on defined grid size of the map |
| **Time-dependent A*[32]** | It can predict human trajectory | It does not concern about obstacle motion |

# 3 Reinforcement Learning in Path Planning

Reinforcement learning (RL) is an effective machine-learning method that uses the results of past actions to reinforce or weaken such actions depending on their success or failure. In mobile robotics, this method uses environmental feedback as the input for path planning. It outputs an action for the robot through continuously interacting with the external environment. With the reinforcement learning mechanism, the robot tries to take actions and receives feedbacks, and then makes decisions based on the feedbacks. Particularly, with the right action, the algorithm will give the robot a positive reinforcement value, while with the wrong action, the algorithm will give the robot a negative value. In the whole process, the robot strengthens its proper behavior and weakens its improper behavior. As a result, a rational solution can be generated for the robot when encountering people and other robots in the environments. Also, as in any type of learning, the performance improves with experience. The process of reinforcement learning is shown in Fig. 6. The policy, which is the path in the navigation process, is generated by interacting with the external environment.

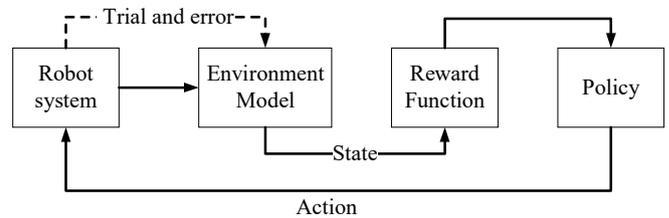

Fig. 6 The process of reinforcement learning

Classical RL algorithms in the robot path planning include the Q-learning algorithm [47], SARSA algorithm [48], R-learning algorithm. Q-learning is among which the most investigated RL algorithm. Commonly, the algorithm outputs a reward value for the robot state and motion, through the feedback that the robot obtains from the environment. Particularly, the Q value of the correct action increases by decreasing the rate of wrong actions. Then, the Q value-based method outputs an optimal policy after the Q value is filtered. The Q-learning algorithm has some limitations as well. First, the memory requirement is large. Second, it will take a long time for learning. Third, the rate of convergence is low. In order to tackle the problems of Q-learning, Peng [49] proposed the Q(λ) algorithm, which utilized the idea of backtracking. In this approach, the subsequent data can be provided back in time so that the method can predict the next behavior in a time-effective manner. The incurring wrong behavior is gradually forgotten in the process of updating.

RL and its variants have been widely used for robot navigation. In order to interact with humans in a graceful manner, robots need to understand and follow certain rules. With that objective, Kuderer [50] proposed an approach to model the cooperative navigation behavior of humans. It is able to obtain human trajectories in real-time. Recently, Inverse Reinforcement Learning (IRL) has gained much research interest. It contains a reward function for the decision-making process. Some researchers applied IRL to get the human comfort model for the collaborative navigation [51]. To achieve graceful path planning in densely populated environments, Chen et al. [52] proposed a decentralized multi-agent collision avoidance algorithm based on deep reinforcement learning, which transfers online computing to offline learning effectively. It is better than the ORCA algorithm strategy can be well extended to new scenarios that do not appear in the training phase. The applications of reinforcement learning in dynamic path planning in recent years are shown in Table 3, in which the advantages and disadvantages of the RL methods are indicated.

## 4 Conclusions

In this paper, the path planning algorithms for autonomous robot navigation were reviewed. The path planning problem under the framework that divides the planning methods into global path planners and local path planners was examined. These methods are effective in solving the while the method may lead to an oscillating (marginally stable) path.

Table 3 The utilization of RL in dynamic motion planning in recent years

| Method | Advantage | Disadvantage |
|---|---|---|
| IRL [51] | Build a human model in different environments, can provide collaborative navigation. | Computational expensive, highly rely on feature selection performance. |
| CADRL [52] | Real-time performance and high path quality | May lead to an oscillating path. |
| SA-CADRL [53][54] | Solve the randomness of human behavior, with respect to common social norms. | Does not consider the relationship between pedestrians. |
| CRL [55] | It can not only solve social-aware motion planning problems but also can interact with human. | Need the prior knowledge. |

SA-CALDRL [53] has been proposed to cope with the randomness of human behavior, in which a time-effective navigation policy is used. Notably, this method can achieve low-speed autonomous navigation of a robot vehicle in a densely populated dynamic environment. However, the algorithm does not take into account the relationship with pedestrians. Everett [54] extended the SA-CALDRL algorithm by introducing the Long Short-Term Memory (LSTM) method in the algorithm. The advantage of the algorithm is that it does not need to assume a specific behavior model. Moreover, it seeks to predict the running direction of the robot in a simple way. In addition, Ciou et al. [55] proposed a Compound Reinforcement Learning (CRL) framework. In this framework, the robot learns graceful social navigation through sensor input. The experiments show that the CRL method can learn to navigate in the environment safely. However, prior knowledge is needed in this framework, limiting its wide application in the real-world. Various environments and various types of stage training framework have been proposed by Long et al. [56]. Such a navigation problem in many applications in different aspects. However, there remains room for further improvement. The progress of reinforcement learning-based path planning methods was explored in the paper, and their capabilities for navigation in complex environments were indicated. The reinforcement learning methods, in the long run, may be encoded into a hierarchical path planning framework.